\documentclass{article}

\usepackage{microtype}
\usepackage{graphicx}
\usepackage{subfigure}
\usepackage{booktabs} 

\usepackage{hyperref}

\usepackage[accepted]{icml2025}

\usepackage{amsmath}
\usepackage{amssymb}
\usepackage{mathtools}
\usepackage{amsthm}

\usepackage[capitalize,noabbrev]{cleveref}

\theoremstyle{plain}

\theoremstyle{definition}

\theoremstyle{remark}

\usepackage[textsize=tiny]{todonotes}

\usepackage{color, colortbl}
\definecolor{Gray}{gray}{0.93}
\definecolor{Orange}{rgb}{1,0.5,0}
\definecolor{DGray}{gray}{0.83}
\definecolor{LightCyan}{rgb}{0.88,1,1}
\usepackage[most, breakable]{tcolorbox}

\usepackage{bbm}
\usepackage{adjustbox}
\usepackage{lscape} 
\usepackage{makecell} 
\usepackage{bm}
\usepackage{float}
\usepackage{xspace}
\usepackage{nicematrix}
\usepackage{url}            
\usepackage{amsfonts}       
\usepackage{nicefrac}       
\usepackage{microtype}      
\usepackage{xcolor}         
\usepackage{pifont}

\usepackage{wrapfig}
\usepackage{multirow,makecell}
\usepackage{tabularx}
\usepackage{bbding}
\usepackage{threeparttable}
\usepackage{wasysym}
\usepackage{todonotes}
\usepackage{enumitem}

\newcommand{\alg}{\textsc{IFPruning}}

\icmltitlerunning{Instruction-Following Pruning for Large Language Models}

\begin{document}

\twocolumn[
\icmltitle{Instruction-Following Pruning for Large Language Models}



\icmlsetsymbol{intern}{$\star$}

\begin{icmlauthorlist}
\icmlauthor{Bairu Hou}{yyy,xxx,intern}
\icmlauthor{Qibin Chen}{yyy}
\icmlauthor{Jianyu Wang}{yyy}
\icmlauthor{Guoli Yin}{yyy}
\icmlauthor{Chong Wang}{yyy}
\icmlauthor{Nan Du}{yyy}
\\
\icmlauthor{Ruoming Pang}{yyy}
\icmlauthor{Shiyu Chang}{xxx}
\icmlauthor{Tao Lei}{yyy}
\end{icmlauthorlist}

\icmlaffiliation{xxx}{UC Santa Barbara}
\icmlaffiliation{yyy}{Apple AI/ML}

\icmlcorrespondingauthor{Bairu Hou}{bairu@ucsb.edu}
\icmlcorrespondingauthor{Tao Lei}{tao\_lei2@apple.com}

\icmlkeywords{Machine Learning, ICML}

\vskip 0.3in
]



\printAffiliationsAndNotice{$^\star$Work done while interning at Apple.}

\begin{abstract}
With the rapid scaling of large language models (LLMs), structured pruning has become a widely used technique to learn efficient, smaller models from larger ones, delivering 
superior performance compared to training similarly sized models from scratch.  In this paper, we move beyond the traditional \textit{static pruning} approach of determining a fixed pruning mask for a model, and propose a \textit{dynamic approach} to structured pruning. In our method, the pruning mask is input-dependent and adapts dynamically based on the information described in a user instruction.  
Our approach, termed ``instruction-following pruning'', introduces a sparse mask predictor that takes the user instruction as input and dynamically selects the most relevant model parameters for the given task. To identify and activate effective parameters, we jointly optimize the sparse mask predictor and the LLM, leveraging both instruction-following data and the pre-training corpus.  Experimental results demonstrate the effectiveness of our approach on a wide range of evaluation benchmarks.
For example, our 3B activated model improves over the 3B dense model by 5-8 points of absolute margin on domains such as math and coding, and rivals the performance of a 9B model.
\end{abstract}

\section{Introduction}

Structured pruning techniques have become a widely adopted method for reducing the inference cost of large language models~\citep{wang2020structured,sreenivas2024llm, muralidharan2024compact, meta2024llama}.  These methods typically optimize a binary mask over language model parameters to minimize either language modeling or task-specific loss~\citep{xiasheared, sreenivas2024llm, meta2024llama}. Once the mask is optimized, the resulting mask is fixed, allowing deployment of a smaller, pruned model. However, the fixed nature of the pruned model poses challenges in real-world inference scenarios, where tasks can vary significantly, for instance, coding, mathematics, or domain-specific requirements, each demanding distinct skills and knowledge from the original language model. A static pruned model may struggle to balance inference efficiency with high performance across diverse tasks.  

Given this, we explore a paradigm shift from \textit{static} pruning masks to \textit{dynamic} ones, addressing the central question:
\begin{center}
    \emph{Can LLMs learn to select the most suited parameters \\based on the task description?}
\end{center}
We aim to automatically generate input-specific pruning masks tailored to the tasks described in user prompts. This dynamic, context-aware pruning mechanism enables the language model to perform inference using only the parameters necessary for the task, offering a compelling balance between efficiency and expressivity compared to using a static dense model.
Moreover, because the parameters are selected and fixed,
our method avoids reloading new parameters during the decoding process. This design choice contrasts with other dynamic methods such as contextual sparsity~\citep{liu2023deja, zhou2024sirius} and mixture-of-experts~\cite{lepikhin2020gshard, fedus2022switch, dai2024deepseekmoe}, which load different parameters at each decoding step, leading to significant weight loading costs.
Our design is particularly suited for on-device models (\emph{e.g.} smartphones and laptops), where the inference typically samples a few responses given the same user query (or the same task). In this case, the same activated parameters are selected and cached as a dense model, therefore achieving the same speedup as the static pruning.

\begin{figure*}[!htb]
    \centering
    \includegraphics[width=0.9\textwidth]{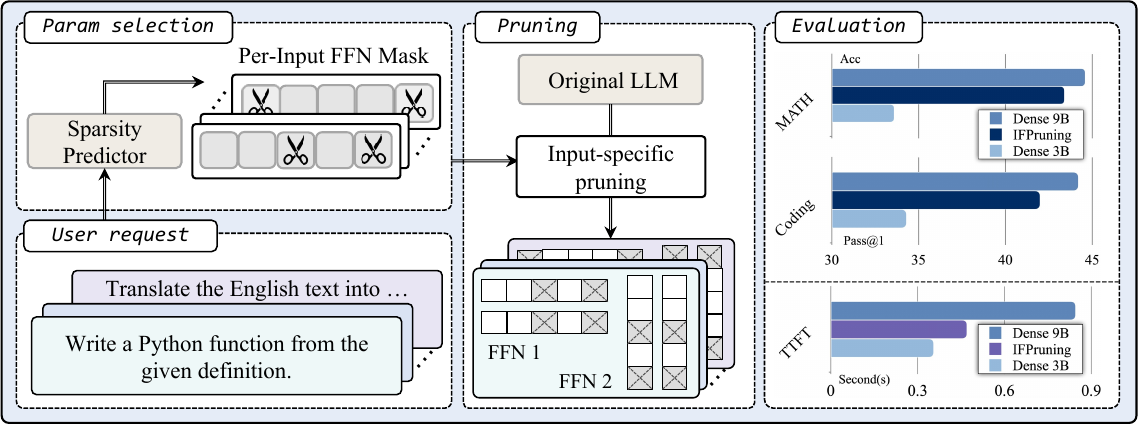}
    \caption{Overview of {\alg}. (Left) For each given prompt, the sparsity predictor (much smaller than the LLM) determines which rows and columns of the FFN matrices should be activated. (Middle) The LLM is then pruned accordingly and uses the selected parameters to perform inference for that specific prompt. (Right) By pruning a 9B LLM to 3B for each input, {\alg} significantly outperforms the dense 3B model and achieves performance levels close to the dense 9B model. It also achieves nearly the same inference latency as the dense 3B model, as measured by time-to-first-token (TTFT).}
    \label{fig: overview}
\end{figure*}

To address the central question, we present \underline{I}nstruction-\underline{F}ollowing \underline{P}runing (\alg), a method that integrates a sparsity predictor with the language model to dynamically generate input-dependent pruning masks, as illustrated in Figure~\ref{fig: overview}. Specifically, we focus on structured pruning of the feed-forward neural network layers, where entire rows or columns of the weight matrices are pruned~\citep{xiasheared, gunter2024apple, sreenivas2024llm}.  
The user prompt is first passed into the sparsity predictor, which assigns importance scores to the rows and columns of each feed-forward network layer.
These scores are then transformed into differentiable masks using the \textsc{SoftTopK} operator~\citep{ainslie2023colt5}, to achieve a predefined number of sparsity (\emph{e.g.}, reducing a 9B language model to 3B active parameters). 
The resulting masks are applied to the language model, in which the feed-forward layers are pruned using the masks.

During training, the differentiable mask generation mechanism allows us to jointly optimize both the sparsity predictor and the language model by minimizing the next-token prediction loss. We employ effective training strategies that leverage both pre-training and supervised fine-tuning data. At test time, only the selected parameters are activated for inference. Parameter selection can be performed either \emph{per-input} or \emph{per-task}: the input prompt can directly be used for parameter selection (Section~\ref{subsec: main_exp}), or a predefined task prompt can be used to select parameters shared across multiple inputs within the same task (Section~\ref{subsec: per_task_pruning}).

We validate {\alg} through comprehensive experiments across diverse tasks in Section~\ref{sec: exp}. Specifically, we fine-tune pre-trained language models of varying sizes (6B, 9B, and 12B parameters) using {\alg} and prune them to activate only 3B parameters.
In particular, {\alg} consistently outperforms 3B dense models across tasks such as math, coding, tool use, MMLU~\cite{hendrycksmeasuring} and AlpacaEval~\citep{dubois2024length}. 
For example, when dynamically pruning the 9B model to 3B, our method improves over the 3B dense model by 8\% on coding tasks and by 5\% on math benchmarks, incurring only marginal performance degradation compared to the unpruned 9B model.

We conduct further analysis to better understand the pruning decisions.
Specifically, we observe that instructions requiring similar skills or domain knowledge yield highly homogeneous pruning patterns.
Inspired by this analysis, we explore per-task pruning in Section~\ref{subsec: per_task_pruning}, where a single task prompt generates shared masks for all test instances within the same task. Results show that per-task pruning maintains robust performance while further reducing data loading overhead.

We also show that {\alg} can significantly improve the LLM inference efficiency in Section~\ref{subsec: latency}. Compared to the full model, {\alg} reduces the time-to-first token by up to 57\% and the generate time by up to 41\%.
In addition, the overhead introduced by dynamic pruning and parameter caching is negligible, adding less than 0.1 seconds per example and accounting for only 1–2\% of the total generation time.

\section{Related Work}
\label{sec: related_work}
In this section, we provide an overview of prior research that closely relates to and partially motivates our work, including model pruning, contextual sparsity, and mixture-of-experts.
\paragraph{Model pruning} Pruning has been extensively studied to compress neural networks and improve their efficiency~\citep{han2015learning, zhu2017prune}. Previous work has explored different pruning techniques for both unstructured pruning ~\citep{narang2017exploring, frankle2018lottery, li2020train, chen2020lottery} and structured pruning~\citep{wen2016learning, voita2019analyzing, louizos2018learning, wang2020structured}. As structured pruning removes entire components in the model such as channels, attention heads,  and feed-forward neural network intermediate dimensions, it is more hardware-friendly than unstructured pruning to compress the large models.

Various methods have been proposed for structured pruning of LLMs~\citep{yang2024laco, kim2024shortened, kurtic2024ziplm, dery2024everybody}. \textsc{LLM-Pruner}~\citep{ma2023llm} adopt the gradient information to find unimportant components in LLMs and remove them. 
\textsc{SliceGPT} transforms each weight matrix in transformer blocks into a smaller one by applying orthogonal transformations to reduce the embedding dimensions of weight matrices.
\textsc{ShortGPT}~\citep{men2024shortgpt} proposes to identify and remove those less important layers, where the layer importance is measured by the similarity between inputs and outputs of that layer.
In comparison, other optimization-based methods directly learn the parameter masks. For example, \textsc{Sheared LLaMA} ~\citep{xiasheared} use the \textsc{HardConcrete} masking~\citep{louizos2018learning, wang2020structured} to generate differentiable masks and optimize the model and masks on pre-training data.
Our method also directly optimize the sparsity predictor and the LLM, and we further extend static pruning to input-dependent pruning.

\paragraph{Contextual sparsity}
Our approach is also directly motivated by the contextual sparsity of LLMs~\cite{liu2023deja, akhauri2024shadowllm, lee2024cats}. Previous work has identified the existence of input-dependent sub-networks (\emph{e.g.}, attention heads and MLP parameters) within ReLU-based LLMs that can generate approximately the same output as the full model for an input. By predicting such sparsity patterns at each decoding step, we can achieve a favorable balance between accuracy and speedup.
But state-of-the-art LLMs~\citep{dubey2024llama, liu2024deepseek, yang2024qwen2} design MLP blocks based on more complex non-linear activation functions such as 
SwiGLU~\citep{shazeer2020glu}, SiLU~\citep{elfwing2018sigmoid, ramachandran2017searching} and GELU~\citep{hendrycks2016gaussian}
that do not inherently induce sparsity~\citep{mirzadeh2023relu, song2024prosparse}. Therefore, directly predicting the sparsity patterns can lead to significant performance degradation~\citep{zhou2024sirius, dong2024prompt}.
In comparison, we co-optimize the sparsity predictor and the LLM with non-ReLU activation functions to achieve better contextual sparsity with minimum performance degradation.
Also, most contextual sparsity methods require predicting sparsity and loading different parameters at each decoding step. Our method eliminates this overhead by selecting the parameters based on the input or task description before decoding starts. The selected parameters are fixed for the entire decoding process, avoiding the parameter reloading cost.

\paragraph{Mixture-of-experts} Mixture-of-Experts (MoE) have emerged as a popular architecture for scaling LLMs while managing inference costs~\citep{lepikhin2020gshard, du2022glam, fedus2022switch, zhou2022mixture, dai2024deepseekmoe, liu2024grin}. These models organize every FFN layer into multiple large FFN blocks referred to as experts, and selectively activate a few experts for each input token via a routing mechanism~\cite{lepikhin2020gshard, zoph2022st, sun2024ec}.
Our method share the same spirit as MoE by dynamically activating a subset of parameters. 
However, our method selects the activated parameters given the input prompt, and reuses the same activated parameters during decoding. 
Although this choice loses the flexibility of using different parameters per token, it significantly reduces weight loading costs for decoding. In this regard, our model is a sparse model designed for on-device scenarios where both memory and computational resources are constrained.
Another difference is that our method performs more fine-grained selection of parameters by activating or pruning each FFN dimension independently, which enhances model expressivity.

\section{Method}
In this section, we elaborate on the details of {\alg}, including the architecture design, data mixture, and training method. We focus on pruning the feed forward blocks (FFNs) in this work, but our method can be easily extend to pruning other components such as attention heads.

\subsection{Overview of Structured Pruning}
Denote the hidden dimension of the LLM as $d$, the intermediate dimension of the FFN blocks as $d_{\mathrm{ffn}}$, the input length as $n$, and $X\in \mathbb{R}^{n\times d}$ as the input of a transformer FFN block $F_{\mathrm{ffn}}(\cdot)$. The goal of our structured pruning method is to reduce the FFN intermediate dimension from $d_{\mathrm{ffn}}$ to $t_{\mathrm{ffn}}$.
Without loss of generality, consider a standard FFN block defined as
\begin{equation}
    F_{\mathrm{ffn}}(X) = \mathrm{FF}_{2}(\mathrm{FF}_{1}(X)) = \sigma(XW_{1})W_2,
\end{equation}
where $W_{1}\in \mathbb{R}^{d \times d_{\mathrm{ffn}}}$, $W_{2}\in \mathbb{R}^{d_{\mathrm{ffn}} \times d}$ are weight matrices, and $\sigma$ is the non-linear activation function.
The structured pruning of the FFN block
can be expressed as applying a mask variable $\bm m \in \{0, 1\}^{d_{\mathrm{ffn}}}$ to the output of the first linear transformation
\begin{equation}
    F_{\mathrm{ffn}}(X, \bm m) = \mathrm{FF}_{2}(\mathrm{FF}_{1}(X)\odot \bm m).
\end{equation}
where $\odot$ is an element-wise multiplication between $\bm m$ and each row of $\mathrm{FF}_{1}(X)$.
For each dimension of $\bm m$, $m_{i} = 0$ indicates that the $i$-th column of $W_1$ and $i$-th row of $W_2$ are pruned. This is because the output $F_{\mathrm{ffn}}(X, \bm m)$ is equivalent to the output of the FFN layer after we prune the $i$-th column of $W_1$ and $i$-th row of $W_2$.
Here $\bm m$ satisfies the sparsity constraint $\sum_{i}m_i = t_{\mathrm{ffn}}$, where $t_{\mathrm{ffn}}$ is the target intermediate dimension of the FFN blocks after pruning.

\subsection{Architecture}
\label{subsec: arch}
As shown in Figure~\ref{fig: overview}, our architecture comprises two key components: a sparsity predictor and a dense LLM to be dynamically pruned. For any given user prompt, the sparsity predictor generates masks that are applied to the LLM backbone, pruning the corresponding rows and columns of the FFN blocks.

\paragraph{Sparsity predictor}
The sparsity predictor consists of two modules: \ding{182} a much smaller LLM backbone to extract the features of user prompts and \ding{183} a mask prediction head.
Specifically, the LLM backbone takes the prompt $\bm{x} = (x_1, \dots, x_n)$ as input with length $n$, and we use the hidden states of the last token $x_{n}$ in the last layer to represent the prompt.
The mask prediction head is a two-layer MLP, which predicts the masks given the prompt presentations. The output of the FFN mask prediction head is the masking score $\bm z \in \mathbb{R}^{L\times d_{\mathrm{ffn}}}$, where $L$ is the number of layers of the LLM.
We include more details about the architecture of the sparsity predictor in Appendix~\ref{app: model_arch}.

Given the predicted masking score $\bm z$, a mask generation operator will be applied to $\bm z$ to convert it to the mask $\bm m \in [0,1]^{L\times d_{\mathrm{ffn}}}$, which contains $t_{\mathrm{ffn}}$ nonzero elements. In this paper, we use the SoftTopK~\citep{lei2023conditional, ainslie2023colt5} algorithm to generate a differentiable $\bm m$, but we also acknowledge that other algorithms such as the HardConcrete masking~\cite{louizos2018learning, wang2020structured} are also applicable. 
Particularly, given the FFN masking score $\bm z$, SoftTopK converts it to masks $\bm m$ via:
\begin{equation}
    \begin{aligned}
        \bm \lambda^{(i)} = g(\bm z^{(i)}),\,\, \bm m^{(i)} = \bm \lambda^{(i)} \odot \mathrm{Top}(\bm \lambda^{(i)}, t_{\mathrm{ffn}}).
    \end{aligned}
\end{equation}
Here $\bm z^{(i)}$, $\bm \lambda^{(i)}$ and $\bm m^{(i)}$ represent the $i$-th row of each matrix, $g(\cdot): \mathbb{R}^{d_{\mathrm{ffn}}} \rightarrow [0, 1]^{d_{\mathrm{ffn}}}$ is a normalization function, and $\mathrm{Top}(\cdot, t_{\mathrm{ffn}}) \in \{0, 1\}^{d_{\mathrm{ffn}}}$ is an indicator function that returns a binary mask indicating the top-k values in $\lambda$.
The normalization function $g(\cdot)$ ensures that $\lambda$ satisfies the sparsity constraint, \emph{i.e.}, $\sum_{k} \lambda^{(i)}_k = t_{\mathrm{ffn}}$, where $t_{\mathrm{ffn}}$ is the target size of the FFN layers. More details of SoftTopK can be found in the previous work~\cite{lei2023conditional, ainslie2023colt5}.

\paragraph{Masked LLM}
During training, the LLM takes the masks $\bm m$ as an additional input and prune its FFN blocks.
We use standard next token prediction loss computed over tokens within a training batch, and we co-optimize the LLM and the sparsity predictor.

\subsection{Model Training}
The training of {\alg} incorporates two stages. We first perform continued pre-training in which we initialize our model using a pretrained dense model, and then perform supervised fine-tuning (SFT) on instruction-following data. In what follows, we elaborate on the details of the two training stages.

\paragraph{Continued pre-training}
\label{subsec: continued_pretrain}
Learning to select input-specific sub-networks may require a lot of training data. Instead of directly training the models on the SFT data only, we first use pre-training data to jointly optimize the sparsity predictor and masked LLM.
Specifically, denoting the input text as $\bm x = (x_1, \dots, x_n)$, we split it into $K$ consecutive chunks with fixed size:
\begin{equation}
    \bm x^{(k)} = x_{(k-1)s+1}, \dots, x_{ks},\,\, k = 1, \dots, K,
\end{equation}
where $s = n/K$ is the fixed size of each chunk.
We then use the each chunk to select parameters of the LLM for the next token predictions in the next chunk, \emph{i.e.},
\begin{equation}
    \begin{aligned}
        \mathcal{L} = \sum_{k=1}^{K-1} \sum_{x_i \in \bm x^{(k+1)}}\ell\left[f(\bm x_{<i};\bm \theta, \bm m^{(k)}), x_i\right],
    \end{aligned}
\end{equation}
where $\bm \theta$ refers to the parameters of the LLM, $\bm m^{(k)}$ is the predicted mask based on chunk $\bm x^{(k)}$, $f(\bm x_{<i};\bm \theta, \bm m^{(k)})$ is the next token prediction distribution from the LLM with $\bm m^{(k)}$ applied, and $\ell(\cdot)$ is the Cross-Entropy loss.
Since the chunks are consecutive, the sparsity predictor can learn to utilize the contextual information of each chunk to predict which parameters of the LLM are best suited for the next token prediction in the next text chunk. 
Because both the sparsity predictor and LLM are co-optimized in this stage, it provides a good initialization for the fine-tuning stage.

\paragraph{Supervised fine-tuning}
Starting from the models after the first stage, we train the sparsity predictor and the LLM on a supervised fine-tuning dataset that contains several million examples.
During training, the input prompts will be fed into the sparsity predictor which predicts the masks for each input. The LLM is then masked and optimized to predict the target outputs conditioned on the input prompts.
Our SFT data contains a diverse set of prompts to predict the sub-networks. Some prompts specify the task with a task description and an input, while others include few-shot examples along with the input. Lastly, many examples only contain a task description, like ``write a comprehensive blog post about the top 10 most eco-friendly cities in the world.''.

For multi-turn conversational data, we only use the first human message as the prompt for sub-network selection.
During training, with the exception of removing all instances of personal data, all these prompts are fed directly into the sparsity predictor without any additional processing. This approach maximizes flexibility during inference, allowing the predictor to generate a sub-network regardless of the prompt format during inference. The training objective follows the standard SFT approach, namely minimizing the cross-entropy loss on the target outputs. Through this process, the model learns to selectively activate the most suited parameters for different input examples.

\section{Experiment}
\label{sec: exp}
In this section, we conduct empirical evaluations to assess the effectiveness of our proposed method.

\subsection{Experiment Setup}
\paragraph{Dataset and backbone models} Our models are trained on an internal SFT dataset with several million examples. We also follow the setup used in \textsc{T\"ulu}\xspace~2 and sample additional 800K examples from the FLAN-V2 collection~\citep{chung2024scaling} to enhance task prompt diversity.
The experiments are conducted using a series of pre-trained LLMs. Particularly, for the sparsity prediction component, we initialize it with a 302M model that has been pre-trained on web-crawled data.
To test the performance of our method across various model scales, we separately train three different models that use 6B, 9B, and 12B parameters, respectively, in the masked LLM component. 
For all these models, our approach activates 3B parameters (and prunes the rest of the parameters).
More details about the model architecture are given in Appendix~\ref{app: model_arch}.

\paragraph{Comparison baselines}
We compare our method with the following models:
\ding{182} \textsc{Dense-3B}. We compare with a 3B dense LLM that is trained using twice as many pretraining tokens compared to our models.
\ding{183} \textsc{Pruning+Distill}.
We also compare our method with static pruning approaches. Similar to recent work such as Sheared LLaMA~\citep{xiasheared}, LLAMA 3.2~\citep{meta2024llama}, and \textsc{Minitron}\citep{sreenivas2024llm}, we include a baseline where a 3B dense LLM is pruned and distilled from a larger pre-trained LLM. we first prune the larger LLM into a dense 3B model by learning masks on the FFN layers similar to Sheared LLaMA~\citep{xiasheared}. After pruning, the model undergoes further continuous pre-training through knowledge distillation~\citep{hinton2015distilling}, using a larger dense model with 12B parameters as the teacher model. Therefore, this approach serves as a stronger baseline compared to models using pruning alone.
More details of the knowledge distillation can be found in Appendix~\ref{app: knowledge_distill}.
\ding{184} \textsc{Dense-9B}. We also include a 9B dense model without pruning as a reference for upper-bound performance.

\definecolor{llamacolor}{HTML}{C6E7FF}
\begin{table*}[ht]
\centering
\caption{Performance comparison between {\alg} and other dense LLMs.
\textsc{Pruning+Distill 3B} is first pruned from larger model into 3B parameters and undergoes continuous pre-training with knowledge distillation using a 12B LLM as the teacher model.
Three versions of {\alg} are included which activate 3B parameters of LLMs with 6B, 9B, and 12B parameters.
The \textbf{best} results are highlighted in \textbf{bold} and the \underline{second-best} results are \underline{underlined}. \textsc{Dense-9B} is included for reference.}
\vspace{0.1in}
\begin{adjustbox}{max width=.88\textwidth}
\begin{tabular}{c p{3cm}|cc|ccc|c}
\toprule
\multirow{2}{*}{\textbf{Category}} & \multirow{2}{*}{\textbf{\,\,\,\,Dataset}} &  \multirow{2}{*}{\makecell{\textsc{Dense} 3B}} &  \multirow{2}{*}{\makecell{\textsc{Pruning}+\\\textsc{Distill} 3B}} & \multicolumn{3}{c|}{\cellcolor{llamacolor}{\alg}} & \multirow{2}{*}{\makecell{\textsc{Dense} 9B}}\\
& & & & \cellcolor{llamacolor} 6B$\rightarrow$3B & \cellcolor{llamacolor} 9B$\rightarrow$3B & \cellcolor{llamacolor} 12B$\rightarrow$3B & \\
\midrule
\multirow{4}{*}{\makecell{\textbf{Instruction}\\ \textbf{Following}}}
& {IFEval-Instruction}    & 85.0 & \textbf{86.7} & 83.9 & \underline{85.9} & 85.3 & 87.5 \\
& {IFEval-Prompt}    & 77.8 & \textbf{80.8} & 77.6 & \underline{78.9} & 78.6 & 81.7 \\
& {AlpacaEval 2.0}  & 27.3 & 30.0 & 29.0 & \underline{31.3} & \textbf{32.5} & 38.6 \\
& {Arena-Hard-Auto}  & 15.8 & 16.4 & 18.0 & \underline{18.6} & \textbf{19.8} & 24.8 \\
\midrule
\multirow{4}{*}{\textbf{Coding}}
& {HumanEval}   & 35.2 & 37.1 & 41.0 & \underline{42.4} & \textbf{43.3} & 46.5 \\
& {MultiPL-E}   & 39.0 & 37.9 & 37.6 & \underline{41.8} & \textbf{43.0} & 44.0 \\
& {MBPP}        & 28.8 & 38.0 & 37.4 & \underline{41.8} & \textbf{42.8} & 42.2 \\
& {Average}     & 34.3 & 37.7 & 38.7 & \underline{42.0} & \textbf{43.0} & 44.2 \\
\midrule
\multirow{2}{*}{\textbf{Tool Use}}
& {Tool Execution} & 74.8 & 74.4 & \textbf{75.8} & \underline{75.7} & 73.9 & 76.5 \\
& {Tool Planning} & 25.2 & 17.5 & 36.6 & \textbf{45.4} & \underline{37.5} & 46.5 \\
\midrule
\multirow{2}{*}{\textbf{Math}}
& GSM8K         & 69.3 & 70.0 & \textbf{72.2} & \underline{72.0} & 70.2 & 75.4 \\
& {MATH}        & 31.8 & 32.7 & 36.2 & \underline{36.7} & \textbf{37.1} & 37.3 \\
\midrule
\multirow{8}{*}{\makecell{\textbf{Core}\\ \textbf{Text}}}
& {ARC-Challenge}       & 47.4 & 46.2 & \underline{50.4} & \underline{50.4} & \textbf{51.9} & 53.9 \\
& {ARC-Easy}       & 79.3 & 79.9 & \underline{81.9} & 81.4 & \textbf{82.3} & 83.4 \\
& {HellaSwag}   & 53.0 & 53.0 & 54.1 & \underline{55.5} & \textbf{55.8} & 57.7 \\
& {LAMBDA}      & 66.5 & 68.2 & 68.8 & \underline{68.9} & \textbf{69.0} & 70.8 \\
& {PiQA}        & 77.4 & 77.3 & 77.7 & \underline{78.0} & \textbf{78.7} & 79.4 \\
& {SciQ}        & 95.9 & 96.0 & 96.4 & \textbf{96.7} & \underline{96.5} & 96.9 \\
& {WinoGrande}  & \textbf{69.8} & \underline{69.1} & 67.7 & 67.0 & 68.4 & 74.3 \\
& {Average}     & 69.9 & 70.0 & 71.0 & \underline{71.1} & \textbf{71.8} & 73.8 \\
\midrule
\textbf{MMLU} 
& {MMLU}        & 61.8 & 62.8 & 63.1 & \underline{65.5} & \textbf{66.1} & 67.8 \\

\bottomrule
\end{tabular}
\end{adjustbox}
\label{tab: main_exp}
\end{table*}

\paragraph{Implementation} We use the AXLearn~\citep{apple2023axlearn} framework and JAX~\citep{bradbury2018jax} for model training. 
Following the previous work~\citep{dubey2024llama}, all the pre-trained model used in our experiments are achieved by performing two-stage pre-training. All models are pre-trained with a batch size of 2048 and a total number of 5T tokens, except that the \textsc{Dense-3B} is trained for 9T tokens.
The SFT training for the baselines and our method is performed with a batch size of 1024 for 60k training steps.
We use the same pre-train and SFT data mixture for all models.

\paragraph{Evaluation configurations} We include the following tasks for evaluation:
\begin{itemize}[leftmargin=10pt,itemsep=2pt, topsep=2pt]
    \item Instruction-following. We include \texttt{IFEval}~\citep{zhou2023instruction}, \texttt{AlpacaEval 2.0}~\citep{dubois2024length}, and \texttt{Arena-Hard-Auto}~\citep{li2024crowdsourced} for evaluation. We report the prompt-level and instruction-level accuracy on IFEval, the length-controlled win rate on AlpacaEval 2.0, and win rate on Arena-Hard-Auto.
    \item Coding tasks. We evaluate the pass@1 performance on \texttt{HumanEval-python}~\citep{chen2021codex}, \texttt{mbpp}~\citep{austin2021program}, and \texttt{MultiPL-E}~\citep{cassano2022multipl}. For MultiPL-E benchmark, we use the Swift subset for evaluation.
    \item Math. We use \texttt{GSM8K}~\citep{cobbe2021training} and \texttt{MATH}~\citep{hendrycksmath2021} to evaluate the math capabilities of LLMs. we report the accuracy with few-shot examples on both datasets (8-shot for GSM8K and 4-shot for MATH).
    \item Core Text. We include a set of tasks to evaluate the model's core capabilities of natural language understanding, scientific knowledge, and reasoning.
     We report the zero-shot performance on \texttt{ARC-challenge}~\citep{clark2018think}, \texttt{ARC-easy}~\citep{clark2018think}, \texttt{HellaSwag}~\citep{zellers2019hellaswag}, \texttt{WinoGrande}~\citep{sakaguchi2021winogrande}, \texttt{PiQA}~\citep{Bisk2020}, \texttt{LAMBADA-OpenAI}~\citep{paperno2016lambada},  and \texttt{SciQ}~\citep{welbl2017crowdsourcing}.
    \item MMLU~\citep{hendrycksmeasuring}. We evaluate the 5-shot performance and report the multiple-choice accuracy.
    \item Tool use. We evaluate the tool use performance on \texttt{MMAU}~\citep{yin2024mmau} and report the performance on Tool Execution and Tool Planning.
\end{itemize}
We mainly use \texttt{LM-Evaluation-Harness}~\citep{gao2021framework} to evaluate the tasks, with the exception of instruction-following and tool-use tasks, which are based on their official implementations.

\begin{figure*}[t]
    \centering
    \includegraphics[width=\textwidth]{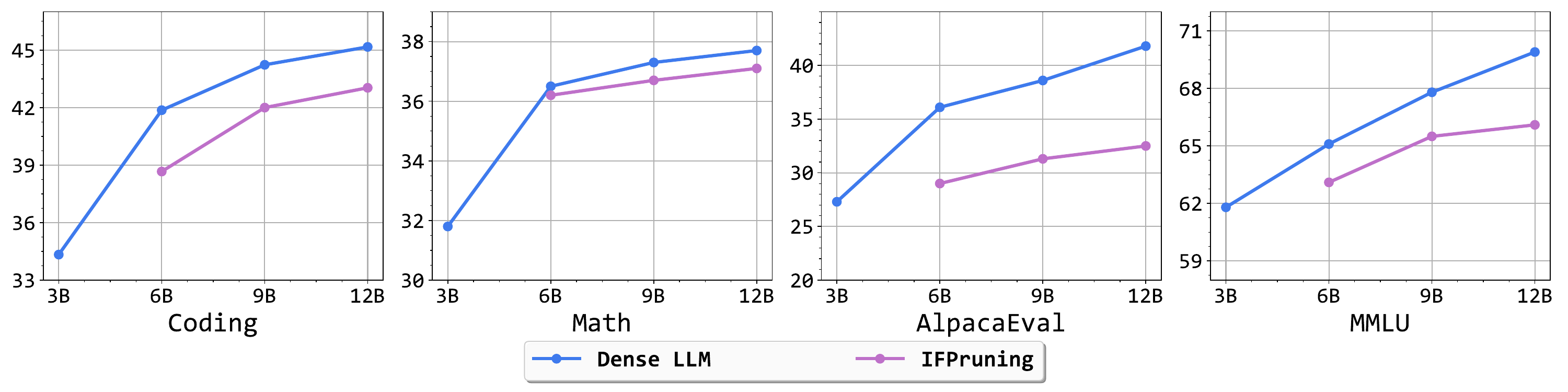}
    \vspace{-5mm}
    \caption{Scaling behavior of dense models and {\alg}. {\alg} activates 3B parameters for each input. The x-axis represents the total number of LLM parameters for dense models and {\alg}, while the y-axis indicates the performance scores.}
    \label{fig: scaling}
\end{figure*}

\begin{figure*}[t]
    \centering
    \includegraphics[width=\linewidth]{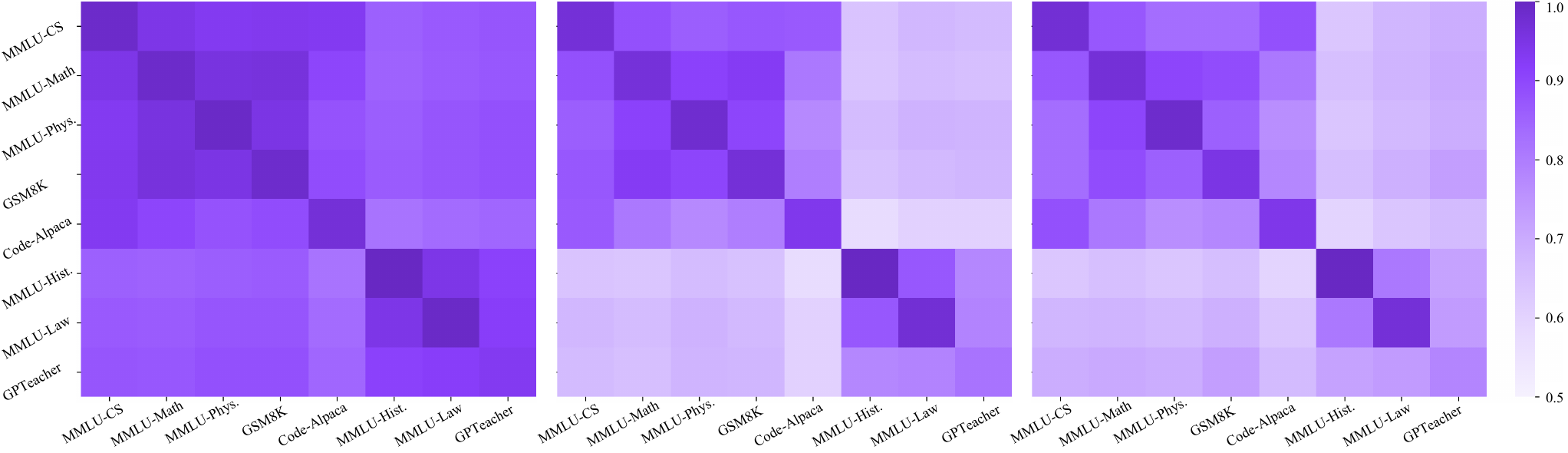}
    \vspace{-5mm}
    \caption{Sub-network overlap rates for the first layer (left), layer 16 (middle), and the last layer (right) of the LLM.}
    \label{fig: overlap_rates}
    \vspace{-2mm}
\end{figure*}

\subsection{Evaluation with Input-Specific Masks}
\label{subsec: main_exp}
In this section, we evaluate our model using input-specific masks, which align with the training scheme. For all examples across the included datasets, we generate masks by feeding the testing question and few-shot examples (if applicable) into the sparsity predictor. The LLM is then pruned with the resulting mask and performs inference on the same input. The datasets provide a diverse range of inputs for the sparsity predictor, including combinations of few-shot examples and testing questions (\emph{e.g.}, MATH and MMLU) and question-only formats (\emph{e.g.}, AlpacaEval and IFEval).

\vspace{-2mm}
\paragraph{Overall comparison} We visualize the evaluation results in Table~\ref{tab: main_exp}. We highlight the following observations.
First, with an equivalent number of activated parameters, {\alg} significantly outperforms the dense LLM, demonstrating its ability to select the most relevant parameters for various inputs effectively. Specifically, {\alg} achieves a 5\% and 4\% higher win rate over the dense LLM on AlpacaEval and Arena Hard, respectively. Also, our method improves upon the dense baseline by 6\% to 14\% on coding tasks and by 3\% to 5\% on math benchmarks. We also observe substantial improvement on Tool Use and the MMLU benchmark.
Finally, {\alg} shows strong performance on Core Text tasks, highlighting its broad applicability. The effectiveness of our approach is further underscored by its performance relative to the 9B-parameter ``upper bound'' model. Notably, on coding, math, and MMLU benchmarks, our method closely approaches upper-bound performance.

Second, {\alg} demonstrates superior performance compared to the structured pruning method. Please note that the structured pruning model, \textsc{Pruning+Distill}, benefits from additional training signals due to knowledge distillation from a larger teacher model as the teacher. 
Nevertheless, {\alg} consistently outperforms this baseline. Our method achieves higher performance across a variety of benchmarks, including AlpacaEval, Arena Hard, math problems, coding tasks, tool use, MMLU, and most Core Text tasks.

Third, we observe a clear improvement in performance with {\alg} as the size of the LLM increases. As the source model size scales from 6B to 9B and then to 12B, there is a noticeable performance boost on most of the datasets.

\vspace{-2mm}
\paragraph{Scaling behavior of dense models and {\alg}}
We illustrate the scaling behavior of dense LLMs (without pruning) and our {\alg} method in relation to model size (total number of parameters) in Figure~\ref{fig: scaling}. In our approach, we consistently activate 3B parameters across LLMs with varying numbers of total parameters.
In general, increasing the size of the LLM leads to performance improvements. This trend is especially clear on Math, coding, and MMLU tasks, where {\alg} achieves performance levels close to the upper bound. 
In contrast, we observe less performance gain on the AlpacaEval dataset, suggesting an opportunity for improvement and/or better understanding of the scaling behavior in future work.

\paragraph{Interpretability of parameter selection} In this section, we examine the parameter pruning and selection patterns across different domains and visualize the similarities between them.
We measure the similarity between two pruned models based on their \textit{overlap rate}, defined as the proportion of parameters commonly activated by both models.
More details of the implementation can be found in Appendix \ref{app: overlap}.
We perform this analysis of {\alg} using our 6B$\rightarrow$3B model. We include the following datasets as our testing domains. For math, we use GSM8K and the college mathematics subset from MMLU (denoted as MMLU-Math). For computer science, we use the college computer science subset from MMLU (denoted as MMLU-CS) and Code-Alpaca~\citep{codealpaca}. We also include the college physics (MMLU-Phys.), high school European history (MMLU-Hist.), and international law (MMLU-Law) subsets from MMLU. For general instructions, we use the GPTeacher~\citep{gpteacher} dataset that is not included in our training data.
The overlap rates for the first, the last, and a middle layer (layer 16) of the pruned models are visualized in Figure~\ref{fig: overlap_rates}.

\begin{table*}[tt]
\centering
\caption{
Performance comparison between {\alg} with per-input masks and per-task masks.
In the per-input setting, {\alg} prunes the LLMs and activates different sub-networks for each input. In the per-task setting, {\alg} selects a single sub-network for all inputs within the same dataset based on a human-written task instruction.
}
\vspace{2mm}
\begin{adjustbox}{max width=.9\textwidth}
\begin{tabular}{cl|c|cc|cc|cc}
\toprule
\multirow{2}{*}{\textbf{Category}} & \multirow{2}{*}{\textbf{Dataset}} &  \multirow{2}{*}{\makecell{\textsc{Dense} 3B}} & \multicolumn{2}{c}{6B$\rightarrow$3B} & \multicolumn{2}{c}{9B$\rightarrow$3B} & \multicolumn{2}{c}{12B$\rightarrow$3B} \\
& & & Per-Input & Per-Task & Per-Input & Per-Task & Per-Input & Per-Task \\
\midrule
\multirow{4}{*}{\textbf{Coding}}
& {HumanEval}   & 35.2 & 41.0 & 39.0 & 42.4 & 40.9 & 43.3 & 45.3 \\
& {MultiPL-E}   & 39.0 & 37.6 & 38.5 & 41.8 & 42.8 & 43.0 & 44.2 \\
& {MBPP}        & 28.8 & 37.4 & 39.0 & 41.8 & 38.2 & 42.8 & 40.4 \\
& {Average}     & 34.3 & 38.7 & 38.8 & 42.0 & 40.6 & 43.0 & 43.3 \\
\midrule
\multirow{1}{*}{\textbf{Math}}
& {MATH}        & 31.8 & 36.2 & 36.6 & 36.7 & 36.8 & 37.1 & 37.8 \\
\midrule
\multirow{5}{*}{\textbf{MMLU}}
& MMLU-physics  & 49.6 & 50.8 & 50.7 & 55.5 & 54.2 & 55.7 & 54.4 \\
& MMLU-math     & 43.7 & 42.8 & 41.8 & 44.0 & 45.8 & 44.2 & 45.5 \\
& MMLU-history  & 73.7 & 75.9 & 75.2 & 78.0 & 74.7 & 78.5 & 77.8 \\
& MMLU-health   & 59.7 & 61.9 & 62.0 & 64.4 & 63.4 & 65.1 & 63.7 \\
& MMLU-business & 75.3 & 73.5 & 75.3 & 76.6 & 75.3 & 75.5 & 76.1 \\
& MMLU-economics& 60.0 & 59.7 & 59.2 & 66.8 & 65.6 & 66.3 & 63.0 \\
\midrule
\multirow{9}{*}{\textbf{Translation}}
& EN-DE         & 33.9 & 35.5 & 35.8 & 35.9 & 35.8 & 33.5 & 36.4 \\
& EN-ES         & 26.9 & 27.4 & 27.2 & 27.0 & 27.2 & 26.8 & 27.5 \\
& EN-FR         & 45.2 & 47.2 & 46.9 & 47.0 & 47.0 & 47.6 & 49.3 \\
& EN-IT         & 28.8 & 30.2 & 30.1 & 30.3 & 30.5 & 31.0 & 30.2 \\
& EN-PT         & 46.6 & 47.0 & 47.9 & 47.1 & 47.2 & 47.4 & 48.1 \\
& EN-ZH         & 35.0 & 41.7 & 39.1 & 41.5 & 40.4 & 42.5 & 40.9 \\
& Average       & 36.0 & 38.2 & 37.8 & 38.1 & 38.0 & 38.1 & 38.7 \\

\bottomrule
\end{tabular}
\end{adjustbox}
\label{tab: per_task_exp}
\end{table*}

We highlight the following findings.
First, in the lower layers, especially the first layer, the LLM tends to activate very similar sub-networks for different inputs. As we move to higher layers, the parameter selection becomes more diverse.
Second, as shown in the figure, {\alg} activates distinct sub-networks for different domains in higher layers. For instance, models pruned for MMLU-CS have substantial overlap with those for Code-Alpaca, significantly more than with other domains. Similarly, models for MMLU-Math, MMLU-Physics, and GSM8K share a high proportion of activated parameters, which diverge notably from the activation patterns for MMLU-History, MMLU-Law, and GPTTeacher.
Third, the overlap rate along the diagonal of the heatmap is very high, reflecting the strong similarity between models for inputs within the same domain.
Finally, as expected, GPTTeacher is an instruction-following dataset that covers a wide range of domains, therefore it does not exhibit significant domain-specific characteristics, resulting in a self-overlap rate that is lower than in other domains.
In summary, the activated parameters are interpretable and reveal clear patterns in different domains. It aligns with the high performance of our method, as the LLM dynamically activates the parameters most suitable for each input.

\vspace{-5mm}
\paragraph{Ablation study} We also conduct an ablation study on the impact of continued pre-training in Appendix~\ref{app: ablation}. Incorporating continued pre-training before SFT leads to consistent and substantial improvements across all benchmarks.

\subsection{Evaluation with Task-Specific Masks}
\label{subsec: per_task_pruning}
An additional noteworthy capability of {\alg} is task-specific pruning. While {\alg} can effectively prune a model on a per-input basis, the algorithm also demonstrates the ability to select one sub-network that can be used for different inputs within the same task or domain.
We design the following experiment to demonstrate the performance of our method with task-specific pruning.

Specifically, we evaluate our models on math problems, coding tasks, MMLU, and machine translation tasks from Flores-101~\citep{goyal2022flores}.
For each dataset, we manually design a concise instruction that describes the task. For example, the instruction for the \texttt{HumanEval} dataset can be: ``You are an expert Python programmer. Write the code which should pass the tests.'' The sparsity predictor takes this instruction as input and selects the corresponding sub-network. This sub-network is then applied consistently to process all testing examples from that dataset, and the model performance on the dataset is evaluated based on the predictions made by the sub-network. The task-specific instructions used for each dataset are provided in Table~\ref{tab: full_per_task_prompt} in Appendix~\ref{app: detailed_per_task_prompt}.
For the MMLU dataset, which consist of multiple subsets representing diverse domains, we group similar subsets into broader domains and assign a single instruction to each domain. For example, the subsets ``astronomy,'' ``college physics,'' ``conceptual physics,'' and ``high school physics'' are grouped into the domain ``MMLU-physics,'' and a single instruction is written for this domain. A detailed list of the subsets included in each MMLU domain is provided in Appendix~\ref{app: mmlu_domain_subset}.

The performance is shown in Table~\ref{tab: per_task_exp}.
We compare the task-specific pruning capabilities of our algorithm with the dense LLM with 3B parameters and the standard {\alg} with per-input mask. Our key findings are as follows.
First, {\alg} can generate high-quality task-specific masks without additional training. we observe substantial performance improvements of the task-specific {\alg} over the dense baseline across all datasets. When applying per-task masks, {\alg} still outperforms the dense LLM by 5\%-12\% on coding tasks, 5\%-6\% on MATH, and 1\% - 5\% on MMLU subsets. For translation tasks, {\alg} selects the appropriate sub-networks without additional training, achieving an improvement of 4.8 points in BLEU score on average.
Note that these task-specific masks are directly predicted by the sparsity predictor, requiring no additional fine-tuning of either the sparsity predictor or the LLM for each task. This highlights the ``zero-shot'' pruning capability of our method.
Second, compared to the standard input-specific {\alg}, the task-specific {\alg} exhibit minimal performance degradation across math problems, coding tasks, and MMLU. The performance gaps are mostly under 1\%, indicating the robustness of our method.

\subsection{Inference Latency Evaluation}
\label{subsec: latency}
Given the strong performance of {\alg}, we further analyze and evaluate its efficiency improvement in this section. The inference pipeline of our method consists of three steps: \ding{182} Parameter selection: given an input, the sparsity predictor selects a subset of parameters to activate. \ding{183} Parameter loading: the selected parameters are loaded, and the corresponding sub-network is cached as a dense model. \ding{184} Generation: the pruned model will encode the input (prefill) and generate a response (decode), following the same procedure as standard dense or MoE models. Sub-network selection only involves a forward pass through the sparsity predictor, which is lightweight (300M parameters in our experiments), so the time cost is minimal. For parameter loading, we maintain a small base model and update its weights with the selected parameters. The generation step is identical to standard inference, allowing our method to achieve the same latency as a smaller dense model.

To illustrate the inference speedup, we run the following simulation. Specifically, we evaluate the inference latency of the models used in our main experiments: \textsc{Dense}-9B, \textsc{Dense}-3B, and {\alg} 9B$\rightarrow$ 3B. Additionally, we test our method on an open-source LLM, LLaMA-3.1-8B-Instruct, where the FFN layers are pruned to reduce the model from 8B to 3B parameters, while keeping other components unchanged.
Although the tests are done on GPUs, we used batch size 1 and 4 generations per query, reflecting on-device usage. The input length is set to 4,000 and the generation length is set to 100. We report the time-to-first-token (TTFT) and the decoding time, both measured in seconds. For dense models (8B and 3B), the TTFT consists of pre-filling only. For our method, we break down TTFT into its components: parameter selection, parameter loading, and pre-filling. The evaluation results are in Table~\ref{tab: latency}.

We highlight the following conclusions. First, similar to standard structured pruning, our dynamic pruning scheme introduced can also significantly reduces inference latency for large-scale models. By pruning a 9B or 8B LLM to 3B, the TTFT is decreased by up to 57\% and decoding time is decreased by up to 41\%. In total, {\alg} can achieve up to 1.8x speedup compared to the original dense LLMs.
Second, the overhead introduced by dynamic pruning and parameter caching is negligible. For each input, the parameter selection and loading only take less than 0.1s in total (1-2\% of the total generation time).
Finally, despite dynamic masking, the runtime of {\alg} is on par with static pruning (\textsc{Dense}-3B baseline), while offering input-specific adaptivity and superior accuracy.
The inference latency evaluation has further verified that our method is suitable for practical deployment on edge devices, as it accelerates inference while achieving superior model performance.

Finally, we compare the inference latency of our method and an MoE model with similar number of activated parameters in Appendix~\ref{app: moe_latency}.
We show that for the on-device inference scenario with small inference batch size, the cost of MoE is multiple times higher than a dense model and our method.

\begin{table}[t]
    \centering
    \caption{Inference latency evaluation. The source models are reduced to 3B by pruning the FFN layers.}
    \vspace{0.1in}
\begin{adjustbox}{max width=.45\textwidth}
    \begin{tabular}{c|ccccc}
    \toprule[1pt]
    Model & \makecell{Parameter \\Selection} & \makecell{Parameter\\Loading} & \makecell{Prefill} & TTFT & \makecell{Decode} \\
    \midrule
    \multicolumn{6}{c}{\textit{Our model}} \\
    \midrule
    \textsc{Dense}-9B & - & - & 0.846 & 0.846 & 7.16 \\
    \textsc{Dense}-3B & - & - & 0.356 & 0.356 & 5.68 \\
    {\alg} & 0.070 & 0.043 & 0.358 & 0.471 & 5.58 \\
    \midrule
    \multicolumn{6}{c}{\textit{Llama-3.1-8b}} \\
    \midrule
    \textsc{Dense}-8B & - & - & 0.702 & 0.702 & 5.47 \\
    \textsc{Dense}-3B & - & - & 0.317 & 0.317 & 3.52 \\
    {\alg} & 0.070 & 0.016 & 0.315 & 0.402 & 3.53 \\
    \bottomrule[1pt]
    \end{tabular}
\end{adjustbox}
    \label{tab: latency}
\end{table}

\section{Conclusion}
In this paper, we extend the structured pruning for LLMs with a dynamic scheme, where the LLM is pruned into different sub-networks given the prompts. With a simple architecture and straightforward training process, our method can significantly improve the model performance compared to dense LLMs with the same number of activated parameters.
In the future, we will test our method when pruning other components of LLMs such as attention heads and hidden dimensions.

This work also opens several promising directions for future exploration.
First, we focus on dynamically pruning LLMs based on contextual information. 
While this approach is well suited for models on consumer-facing devices such as phones,
additional challenges remain for server-side serving when the input is a batch of user requests containing different tasks.
One possible solution is to cluster user requests so that requests within the same batch share similar activated sub-networks.
Second, the current training method relies on end-to-end optimization, which may not fully utilize the training examples. The performance could be improved by adopting more advanced training strategies.
For instance, using contrastive loss could encourage higher overlap rates among sub-networks for similar inputs, making the model more robust.

\section*{Acknowledgements}
Shiyu Chang acknowledges support from National Science Foundation (NSF) Grant IIS-2338252, NSF Grant IIS-2207052, and NSF Grant IIS-2302730.

\section*{Impact Statement}
In this paper, our primary goal is to develop an algorithm that can dynamically select the most suited parameters of an LLM given an input prompt. Our method is designed to improve both inference efficiency and the performance of LLMs. The training data has been carefully filtered to ensure quality and safety; for instance, all instances of personal data in the SFT data were removed to uphold privacy standards. All the data collection process strictly adheres to ethical guidelines for data use, ensuring that no private or sensitive information is included in the training or evaluation process.

Also, The sparsity-inducing mechanism proposed in this work does not introduce additional risks of bias or harm with the underlying large language model. Furthermore, our method enhances computational efficiency, potentially reducing the environmental impact of large-scale model inference. While acknowledging that any machine learning model has the potential for misuse, we focus on safe and task-specific applications, such as math, coding, and tool use. We encourage further research into mitigating biases and unintended consequences in language models and remain committed to the responsible and ethical advancement of AI technologies.

\bibliography{example_paper}
\bibliographystyle{icml2025}

\newpage
\appendix
\onecolumn
\section{Appendix}
\label{sec:appendix}

\subsection{Implementation Details}
\label{app: model_arch}
\paragraph{LLM architecture}
The LLMs used in this paper follow standard LLM design~\citep{meta2024llama, qwen2.5}, such as grouped-query attention and RMSNorm, with no custom components.
We list the detailed model architecture of the pre-trained models used in our experiments in Table~\ref{tab: model_arch}. 
All models use the same model dimension, attention dimensions and the number of Transformer layers. The only difference is the feed-forward dimension. Accordingly, our method learns to select 6656 FFN dimensions from the 6B and 9B model.

\begin{table}[h]
\centering
\caption{Detailed Architecture of the pre-trained models used in our experiments}
\vspace{0.1in}
\begin{adjustbox}{max width=.9\linewidth}
\begin{tabular}{l|ccc}
\toprule
                    & 3B & 6B & 9B \\
\midrule
    Model Dimension & 2048 & 2048 & 2048 \\
    FFN Dimension & 6656 & 16384 & 24576 \\
    Head Dimension  & 128  & 128  & 128  \\
    Num query heads & 16   & 16   & 16     \\
    Num key/value heads    & 2   & 2    &  2    \\
    Num layers      & 56   & 56   &  56  \\
\bottomrule
\end{tabular}
\end{adjustbox}
\label{tab: model_arch}
\end{table}

\paragraph{Sparsity predictor} The predictor is built upon a lightweight LLM with 302M parameters, built on a pre-trained LM backbone. It consists of:
\begin{itemize}[leftmargin=10pt,itemsep=2pt, topsep=2pt]
\item A small pre-trained language model with 302M parameters as the feature extractor. Given an input, the last hidden state of the final input token will be used as the features of the input.
\item A Two-layer MLP as the prediction head. Given the extracted feature vector with dimension $d$, the first layer transform the input from $d$ to 128. The second layer transform the output of the first layer into $L * d_{\mathrm{ffn}}$, where $L$ is the number of layers of the source LLM being pruned and $d_{\mathrm{ffn}}$ is the intermediate dimension of the source LLM. Then we reshape the output vector into a $l$-by-$d_{\mathrm{ffn}}$ matrix. Each row of the matrix will be processed by the \textsc{SoftTopK}~\citep{lei2023conditional} operator independently to get the FFN mask for the corresponding layer.

\end{itemize}

\subsection{Training of the Structured Pruning Baseline}
\label{app: knowledge_distill}
Our pruning baseline, \textsc{Pruning+Distill}, first prunes the given LLM into a dense 3B model then perform continuous pre-training through knowledge distillation. For knowledge distillation.
we apply KL divergence between the output distributions of the student and the teacher model for each output token. The teacher distribution only keeps the highest-scoring tokens, similar to Minitron~\citep{muralidharan2024compact}. We minimize a combined loss of the standard next-token prediction and KL divergence loss.

\subsection{Interpretability of Parameter Selection}
\label{app: overlap}
We provide more details of the visualization of the parameter selection pattern in Figure~\ref{fig: overlap_rates}. First, we sample 128 inputs per dataset (MMLU, GSM8K, CodeAlpaca-20K, GPTTeacher). Inputs to the sparsity predictor are formatted with in-context examples (MMLU: 5-shot, GSM8K: 8-shot) or raw prompts (CodeAlpaca, GPTTeacher). After that, each input is passed through the sparsity predictor, which selects a fixed number of FFN units (a binary mask) at each layer, producing 128 sub-networks per dataset.  To calculate the overlap, We compare every pair of sub-networks within the 128 examples. For each pair, we compute the fraction of selected FFN units they share. The final overlap rate is the average of these pairwise overlaps.

\subsection{Ablation Study on the Continued Pre-training}
\label{app: ablation}
In this section, we conduct a ablation study on the impact of the continued pre-training. We apply our method and activate 3B parameters of a \textsc{Dense}-6B LLM. We train two variants of models: one with the SFT phase ony and one with both continued pre-training and SFT phases. The experiment results are shown in Table~\ref{tab: ablation_pretrain}. We observe consistent and notable gains across all benchmarks, demonstrating the effectiveness of the continued pre-training phase.

\begin{table}[t]
    \centering
    \caption{Ablation study on the impact of the continued pre-training. We apply {\alg} to a dense LLM with 6B parameters and activate 3B parameters for each input.}
    \vspace{0.1in}
    \resizebox{0.75\linewidth}{!}{
    \begin{tabular}{l|cccccc}
    \toprule
    & HumanEval & MBPP & MultiPL-E & GSM8K & MATH & MMLU \\
    \midrule
    SFT Only &  25.3 &24.4 &15.3 &50.9 &13.5 &55.2 \\
    Continued pre-training + SFT &  31.9 &35.3 &22.4 &61.3 &20.1 &59.3 \\
    \bottomrule
    \end{tabular}
    }
    \label{tab: ablation_pretrain}
\end{table}

\subsection{Inference Latency Comparison with MoE}
\label{app: moe_latency}

\begin{wrapfigure}{r}{0.45\textwidth}
\centering
\vspace{-8mm}
    \caption{Inference latency evaluation. The source models are reduced to 3B by pruning the FFN layers.}
    \vspace{0.1in}
\begin{adjustbox}{max width=.97\linewidth}
    \begin{tabular}{l|ccccc}
    \toprule[1pt]
    Model & \makecell{Param.\\selection} & \makecell{Param.\\loading} & \makecell{Prefill} & TTFT & \makecell{Decode} \\
    \midrule
    \textsc{Dense}-9B & - & - & 0.846 & 0.846 & 7.16 \\
    \textsc{MoE}-A2.7B & - & - & 0.621 & 0.621 & 28.43 \\
    \textsc{Dense}-3B & - & - & 0.356 & 0.356 & 5.68 \\
    {\alg} & 0.070 & 0.043 & 0.358 & 0.471 & 5.58 \\
    \bottomrule[1pt]
    \end{tabular}
\end{adjustbox}
    \label{tab: moe_latency}
\end{wrapfigure}
In this section, we compare the latency of our method with the open-source model, Qwen1.5-MoE-A2.7B. It has 14.3B parameters in total and activates 2.7B parameters per token. For our method, we prune LLaMA-3-8B to 3B parameters. The experiment configurations are exactly the same as the experiment in Table~\ref{tab: latency}.
Specifically, we evaluate the latency on a single NVIDIA RTX A6000 GPU and report time-to-first-token (TTFT) and decoding time with input length = 4k, generation length = 100, and sample 4 responses for each query. The experiment results are shown in Table~\ref{tab: moe_latency}.

We can see the dense baseline and our method have significantly better latency and throughput than MoE. The main reason is that MoE models are not efficient for on-device inference when generating responses given a single query, where the decoding is bottlenecked by weight loading. Since MoE requires reading many expert weights (e.g., 2 for each token), the cost of MoE is multiple times higher than a dense model and our method. This further demonstrates that our method is more suited for on-device scenarios.

\subsection{Subsets included in MMLU domains}
\label{app: mmlu_domain_subset}
We list the subsets in each MMLU domain for the experiment in Table~\ref{tab: per_task_exp}.

\noindent\textbf{MMLU-physics}: astronomy, college physics, conceptual physics, and high school physics.

\noindent\textbf{MMLU-math}: abstract algebra, college mathematics, elementary mathematic, high school mathematics, and high school statistics.

\noindent\textbf{MMLU-history}: high school european history, high school us history, high school world history, and prehistory.

\noindent\textbf{MMLU-health}: anatomy, clinical knowledge, college medicine, human aging, medical genetics , nutrition, professional medicine, and virology.

\noindent\textbf{MMLU-business}: business ethics, management, and marketing.

\noindent\textbf{MMLU-economics}: econometrics, high school macroeconomics, and high school microeconomics.

\subsection{Detailed Per-task Prompts}
\label{app: detailed_per_task_prompt}
In this section, we list the full prompts for the task-specific pruning in Section~\ref{subsec: per_task_pruning} in Table~\ref{tab: full_per_task_prompt}.

\begin{table*}[t]
\centering
\small
\caption{Task prompt for task-specific pruning experiments.}
\vspace{0.1in}
\resizebox{\textwidth}{!}{
\begin{tabular}{@{}p{0.2\textwidth}p{0.8\textwidth}@{}}
\toprule \midrule
\multicolumn{1}{l}{MATH}
& 
\makecell[l]{
You are a Math expert. You will be given a math problem in domains such as algebra, probability, \\ geometry and number theory. Reason and give a final answer to the problem. Your response\\ should end with ``The answer is [answer]`` where [answer] is the response to the problem.} 
\\ \midrule
\multicolumn{1}{l}{MMLU-Physics}
& 
\makecell[l]{
You are an expert in physics. You will be given multiple choice questions in subjects such as \\ astronomy, conceptual physics, and college physics. Select the correct answer to each question.
} 
\\ \midrule
\multicolumn{1}{l}{MMLU-History}
& 
\makecell[l]{
You are an expert in history. You will be given multiple choice questions in subjects such as \\ european history, us history and prehistory. Select the correct answer to each question.
} 
\\ \midrule
\multicolumn{1}{l}{MMLU-Economics}
& 
\makecell[l]{
You are an expert in economics. You will be given multiple choice questions in subjects such as \\ econometrics, macroeconomics and microeconomics. Select the correct answer to each question.
} 
\\ \midrule
\multicolumn{1}{l}{Translation (EN-DE)}
& 
\makecell[l]{
You are a skilled translator who specializes in English to German translations. Your task is to \\ accurately translate the provided English text into German while preserving the meaning and context.
} 
\\ \midrule
\multicolumn{1}{l}{Translation (others)}
& 
\makecell[l]{
Same as above. Replace the language names with the language pair being tested.
} 
\\ \midrule
\multicolumn{1}{l}{Multiple-E Swift}
& 
\makecell[l]{
You are an expert Swift programmer. You will be given a Swift function definition in documentation \\ comments after ///. Write the code to complete the function.\\ Here is an example input:\\  \texttt{\`{}\`{}\`{}}swift \\ /// Write a swift function to count inversions in an array.\\ func get\_Inv\_Count(arr: [Int]) $\rightarrow$ Int
} 
\\ \midrule
\multicolumn{1}{l}{HumanEval-Python}
& 
\makecell[l]{
You are an expert Python programmer. You will be given a Python function definition and some test\\
examples in triple quotes \texttt{"{}"{}"{}}. Write the code which should pass the tests.\\ Here is an example input:\\  \texttt{\`{}\`{}\`{}}python \\
def greatest\_common\_divisor(a: int, b: int) -> int:\\
\texttt{"{}"{}"{}}Return a greatest common divisor of two integers a and b\\
\texttt{>>>} greatest\_common\_divisor(3, 5)\\
1\\
\texttt{>>>} greatest\_common\_divisor(25, 15)\\
5\\
\texttt{"{}"{}"{}}
} 
\\ \midrule
\multicolumn{1}{l}{Mbpp}
& 
\makecell[l]{
You are an expert Python programmer. You will be given a Python function definition and some test\\
examples in triple quotes \texttt{"{}"{}"{}}. Write the code which should pass the tests.\\ Here is an example input:\\  \texttt{\`{}\`{}\`{}}python \\
\texttt{"{}"{}"{}}\\
Write a function to find the similar elements from the given two tuple lists.\\
assert similar\_elements((3, 4, 5, 6),(5, 7, 4, 10)) == (4, 5)\\
\texttt{"{}"{}"{}}\\
} 
\\
\midrule \bottomrule
\end{tabular} 
}
\label{tab: full_per_task_prompt}
\end{table*}

\subsection{License of Datasets and Use of Generative AI}
\label{appendix: license}
We include a diverse set of datasets for evaluation, and their licenses are detailed below. The IFEval dataset is released under the Apache License 2.0. AlpacaEval 2.0 and Arena Hard are also under the Apache-2.0 License. HumanEval, GSM8K, MATH, HellaSwag, WinoGrande, and LM-Evaluation-Harness are under the MIT License. The mbpp dataset is distributed under the Creative Commons Attribution 4.0 license, while MultiPL-E is under the BSD 3-Clause License with Machine Learning Restriction. The ARC dataset is provided under the Creative Commons Attribution Share Alike 4.0 license, and SciQ is under the Creative Commons Attribution Non-Commercial 3.0 license. PiQA is licensed under the Academic Free License v. 3.0. Additionally, MMLU is under the MIT License, and MMAU is distributed under the Creative Commons Attribution 4.0 license. 

The usage of all datasets and packages in this work aligns with their intended purposes, specifically the evaluation of  LLMs.

We utilized GPT-4 to assist in checking and refining grammar and clarity across all sections. The core ideas, analyses, and textual composition remain entirely the work of the authors.


\end{document}